*Research Paper*

# AI-Based Screening for Depression and Social Anxiety Through Eye Tracking: An Exploratory Study.

Karol Chlasta *
Katarzyna Wisiecka **
Krzysztof Krejtz ***
Izabela Krejtz ****

## ABSTRACT

Well-being is a dynamic construct that evolves over time and fluctuates within individuals, presenting challenges for accurate quantification. Reduced well-being is often linked to depression or anxiety disorders, which are characterised by biases in visual attention towards specific stimuli, such as human faces. This paper introduces a novel approach to AI-assisted screening of affective disorders by analysing visual attention scan paths using convolutional neural networks (CNNs). Data were collected from two studies examining (1) attentional tendencies in individuals diagnosed with major depression and (2) social anxiety. These data were processed using residual CNNs through images generated from eye-gaze patterns. Experimental results, obtained with ResNet architectures, demonstrated an average accuracy of 48% for a three-class system and 62% for a two-class system. Based on these exploratory findings, we propose that this method could be employed in rapid, ecological, and effective mental health screening systems to assess well-being through eye-tracking.

_______

* Kozminski University, Warsaw, Poland. Email: karol@chlasta.pl
** University of Economics and Human Sciences, Warsaw, Poland. Email: k.wisiecka@vizja.pl
*** SWPS University, Warsaw, Poland. Email: kkrejtz@swps.edu.pl
**** SWPS University, Warsaw, Poland. Email: ikrejtz@swps.edu.pl

# 1. INTRODUCTION

Well-being is a dynamic construct that changes over time and fluctuates within individuals, making it difficult to quantify (Sonnentag, 2015). This paper attempts to deepen our understanding of how predictive artificial intelligence (AI) methods can be applied to quantify person's well-being. Although current methods, such as the WHO-5 questionnaire by Topp, Østergaard, Søndergaard, and Bech (2015), help explain respondents' subjective well-being, we believe that future solutions should be more less subjective and more automated. They should be based not only on the self-assessment questionnaires, but also on the data gathered using a mental health screening system, that could leverage a range of digital technologies. This paper proposes a novel approach to quantify person's well-being (or lack of thereof, when depression or social anxiety is present), using the well-established methods like eye tracking and convolutional neural networks (Sonnentag, 2015).

Affective disorders may moderate typical patterns of facial emotional expressions. Several studies found that the natural bias to focus on positive stimuli, typically observed among non-dysphoric individuals, does not occur among dysphoric people (Ellis, Fischer, and Beevers, 2010); (Leyman, De Raedt, Vaeyens, & Philippaerts, 2011); (Sears, Newman, Ference, & Thomas, 2011). We are also typically attracted to those people who smile, and we tend to avoid people who express sadness (Van Kleef, Van Doorn, Heerdink, & Koning, 2011).

Cognitive theories assume that attentional biases to negative emotional stimuli play an important role in the development and maintenance of depression (Armstrong and Olatunji, 2012); (Clark, 1999); (Foland-Ross & Gotlib, 2012); (Williams & Scott, 1988). Eye-tracking studies have found that depression is associated with an increased number of fixations spent on negative information (Caseras, Garner, Bradley, & Mogg, 2007); (Eizenman et al., 2003); (Kellough, Beevers, Ellis, & Wells, 2008), and a decreased amount of time spent looking at positive stimuli (De Raedt & Koster, 2010); (Peckham, McHugh, & Otto, 2010). Dysphoric individuals are also slower in disengaging visual attention from depression-related images (Sears, Thomas, LeHuquet, & Johnson, 2010). Sanchez (2013) Sanchez, Vazquez, Marker, LeMoult, & Joormann (2013) showed that participants with Major Depressive Disorder (MDD) compared to control participants disengaged significantly longer from sad faces. Duque and Vázquez (2015) revealed that compared to non-depressive, participants, those with MDD showed also a negative bias for sad faces in attentional maintenance indices (i.e. first fixation duration and total fixation time). Furthermore, the MDD group spent a marginally less amount of time viewing happy faces compared to the non-depressive group.

These attention biases are usually observed using valenced facial emotional expressions as stimuli (Armstrong & Olatunji, 2012). Tedious processing of social stimuli leads to more persistent negative affect (Disner, Shumake, & Beevers, 2017). Processing of social stimuli has been found to be related to greater rumination in depressed individuals (Donaldson, Lam, & Mathews, 2007); (Owens & Gibb, 2017) and a decrease in self-esteem among socially anxious individuals (Iancu, Bodner, & Ben-Zion, 2015).

Cognitive models of social anxiety (Clark & Wells, 1995); (Rapee & Heimberg, 1997) also underline the role of biased processing of socially threatening information in the development of this disorder. Attentional biases in anxious individuals are guided by two theoretical approaches: the vigilance hypothesis, and/or the maintenance hypothesis. The vigilance hypothesis predicts faster orienting toward threat related stimuli. The maintenance hypothesis suggests difficulty in disengaging attention from threat.

What combines the vigilance and maintenance approach is the emotional withdrawal from processing threatening facial expressions. For example, social phobics have been found to exhibit greater hyperscanning of face stimuli than controls (Horley, Williams, Gonsalvez, & Gordon, 2003, 2004). This hyperscanning strategy, reflected by an increase in scan path length may suggest ambient processing of facial emotional expressions (Krejtz et al., 2018).

The present study examines the automated depression and social anxiety detection in eye movements (scan paths) using deep learning among adults. We expect that eye movements during free viewing of emotional expressions can be a good predictor of these disorders.

## 2. MATERIAL AND METHODS

### 2.1 Participants

The database contains eye-tracking data collected during two studies evaluating: (1) attentional tendencies among individuals diagnosed with depression (Holas, Krejtz, Wisiecka, Rusanowska,

& Nezlek, 2020) and (2) social anxiety (Krejtz et al., 2018). Eligibility for the depression group required a diagnosis of a major depressive episode at the time of the study, determined using the Mini-International Neuropsychiatric Interview (MINI), a standard diagnostic tool for DSM-IV and ICD-10 (Sheehan et al., 1998). For the second, social anxiety group, participants first completed an online version of the Liebowitz Social Anxiety Scale (LSAS), a self-reported measure assessing fear and avoidance of social situations (Liebowitz, 1987). The LSAS has a total score range of 0 to 144, derived from the sum of fear and avoidance scores, with common thresholds for interpretation as follows: <55 indicates mild social anxiety, 55–65 moderate social anxiety, 66–80 marked social anxiety, 81–95 severe social anxiety, and >95 very severe social anxiety. Participants then completed the Centre for Epidemiological Studies-Depression Scale (CES-D), a 20-item inventory designed to measure depressive symptoms (Radloff, 1977). The CES-D has a total score range of 0 to 60, with scores of 16 or higher suggesting possible clinical depression and 27 or higher indicating a high level of depressive symptoms.

Participants were invited to the laboratory for individual eye-tracking sessions based on their scores. Individuals were excluded from both studies if they had a current or lifetime psychotic disorder, bipolar disorder, substance abuse, or current suicidal tendencies. The sample contained 53 depressive participants in total (39 Female, M age = 34.60, SD = 8.38) in the first study and 24 socially anxious (19 Female, M age = 23.68, SD = 7.24) with 24 controls (17 Female, M age = 25.76, SD = 6.64) in the second study, as described in Table 1.

Table 1. Descriptive statistics of depression and social anxiety scales

| Group | CES-D Score | LSAS Score |
|---|---|---|
| Anxious | M = 25.67, SD = 11.12 | M = 78.92, SD = 13.68 |
| Control | M = 15.75, SD = 8.62 | M = 32.83, SD = 13.44 |
| Depressive | M = 43.59, SD = 6.67 | M = 71.06, SD = 23.65 |

Note: M= Mean, SD= Standard Deviation

## 2.2 Experimental Task

Before participating in the experiments, participants signed an informed consent form and completed a 5-point calibration of the eye tracker. After that they were asked to free view a series of 12 slides. Each slide presented four categories of facial expressions: neutral, sad, angry, and happy expressed by the same person. There were 6 female and 6 male faces taken from the Karolinska Directed Emotional Faces database (Goeleven, De Raedt, Leyman, & Verschuere, 2008). Each slide was presented for 10 seconds. Although 10 seconds might seem too long for a free viewing study, a 10-second stimulus display is frequently used in depression research with eye tracker (Armstrong & Olatunji, 2012). The slides were presented in the resolution of 1680×1050 pixels in the viewing distance of 60 cm.

## 2.3 Dataset Description and Pre-processing

The eye movements were recorded with an SMI eye tracker at 120 Hz, as presented in Figure 1. Raw data were processed with SMI's BeGaze software. SMI's BeGaze dispersion-based algorithm was used for detection of fixations and saccades. Fixations were defined as stable eye movements, within 1 degree of visual angle, lasting for at least 80 ms. Fixations of duration 80–1200 ms were analysed as well as saccades of amplitude <10◦.

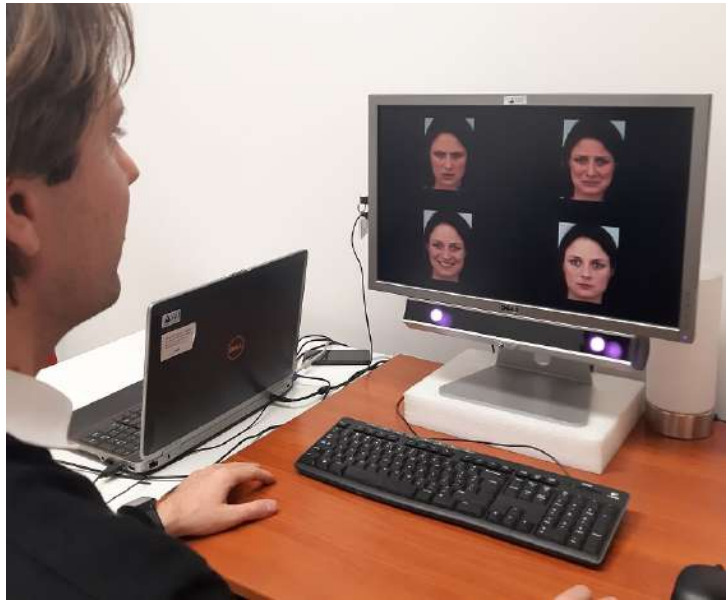

**Figure 1.** Experimental Setting and SMI 120 Hz eye tracker

We selected a single slide from each free viewing session to ensure that the faces seen by participants were displayed. The visualisations of gaze patterns were extracted from BeGaze in a png format. We decided to extract scan paths with calculated fixations in Dataset A1, and raw scan path in Dataset B. In both approaches visualisations of gaze patterns were applied as an input to artificial neural networks.

To balance the number of trials in each group the visualisations chosen for machine learning were selected based on calibration and data quality. In total we selected 60 scan path visualisations in Dataset A1 (scan paths and fixations), and 59 in Dataset B (raw scan paths), as presented in the below Figure 2.

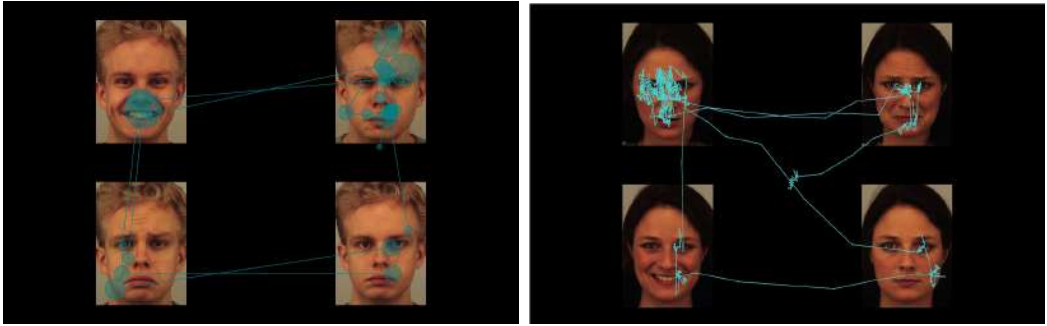

**Figure 2**. Exemplary scan paths of depressive participants before pre-processing

One of our key considerations while applying machine learning algorithms to our problem was limited availability of data, which could cause the deep learning methods to capture the noise, rather than the predictive nature of the dataset. To overcome the problem, we added more data (by applying custom image augmentation using nine ImageMagick filters to create a larger Dataset A2), used batch normalisation, and regularisation in training neural networks (Shang, Chiu, and Sohn, 2017), and explored ResNet architectures with different complexities (from 18 to 151 layers).

Our initial approach was to extract eye-gaze patterns from the images to reduce the amount of noise that could affect the feature selection in the machine learning process. Extracting these patterns for scan paths with fixations drawn on faces (containing a level of opacity) proved to be a manual and time-consuming task with GIMP. Therefore, we decided to augment the data using image filters (Dataset A2), rather than denoise it by removing faces. As a result, we created three datasets A1, A2 and B that were used to train and test our approach. This allowed exploratory evaluation of both the method, and each dataset.

For Datasets A1/A2 the image of each eye-gaze pattern was resized to fit a 224×224 px size. Standard Fastai (Howard and Gugger, 2020) transformations were applied at the time of training to create input for CNN. Additionally, a custom augmentation to increase the size of the dataset

A1 was performed using a command-line ImageMagick (Still, 2006) to create a larger Dataset A2, as presented on Figure 3.

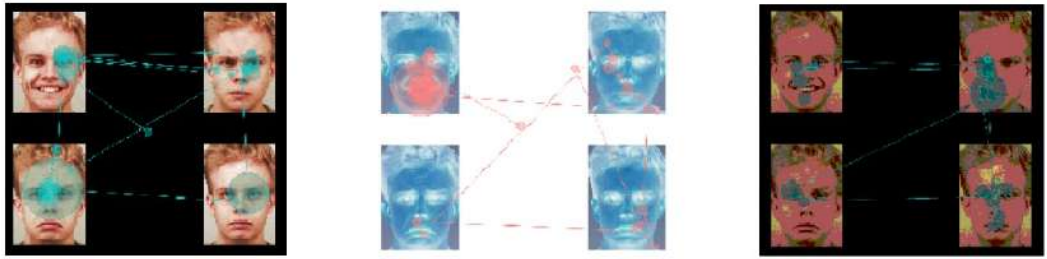

**Figure 3.** Demonstration of exemplary augmentations: gamma, negative, posterize (Dataset A2)

Each image from Dataset A1 was enhanced with the following filters: negate, canny 10, posterize 2, posterize 4, paint 1, paint 3, gamma 100, modulate 140, modulate 160. The filters were selected to retain the eye-gaze pattern in the image and applied on a folder with Dataset A1 images using a bash command1.

For Dataset B, the image files were pre-processed with the GNU Image Manipulation Program (GIMP) (Van Gumster & Shimonski, 2011). We extracted the scan paths from each image, so that our method could identify key characteristics of eye-gaze patterns in the training data, and later match them in the test data. The image of each eye-gaze pattern was resized to fit 448×448 px size and saved in colour with a transparent background to create input for machine learning, as presented in Figure 4.

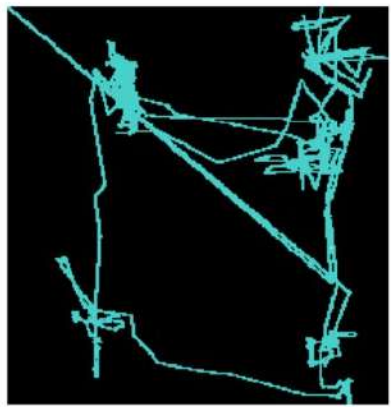

**Figure 4.** Sample image presenting raw scan path generated by an anxious participant (Dataset B)

---

[1] for file in *.png; do convert $file -filter $file; done

Finally, the data were split into training and test sets by randomly assigning 80% of the items to the training set and 20% to the test set.

**Table 2.** Number of samples in Dataset A1, extended Dataset A2 (Fixations) and in Dataset B (Scan paths)

| Category | Dataset A1 | Dataset A2 | Dataset B |
|---|---|---|---|
| Anxious | 20 | 200 | 24 |
| Control | 20 | 200 | 18 |
| Depressive | 20 | 200 | 17 |
| Total | 60 | 600 | 59 |

## 2.4 Method

Our method is based on deep CNNs (Krizhevsky, Sutskever, & Hinton, 2012), and built in Python as PyTorch models (Paszke et al., 2019). We extend the work presented in Chlasta et al. (2021), where a DemCNN model was 63.6% accurate in dementia screening. In the current study, we use graphical representation of all suitable eye-tracking data sourced from 53 depressive participants (39 Female, M age = 34.60, SD = 8.38), 24 socially anxious (19 Female, M age = 23.68, SD = 7.24) with 24 controls (17 Female, M age = 25.76, SD = 6.64).

We use this dataset to explore the transfer learning capabilities of ResNet-18, ResNet-34, ResNet-50, ResNet-101, and ResNet-152 models, which employ a residual learning framework (He, Zhang, Ren, and Sun, 2016). We acknowledge that the use of recurrent networks for sequential data is intuitive; however, (1) the data we gathered were in the form of images, and (2) we anticipated that the predictive capabilities of the dataset would be more closely related to an overall 'map' of gaze patterns for a given task, rather than individual sequences of eye movements. The CNN models we employed conclude with several linear layers. The final convolutional layer extracts features from the image processed by the model and converts them into predictions for each output class. We utilised transfer learning by applying pre-trained weights from ImageNet (Deng et al., 2009) to all convolutional layers, while the final linear layers were randomly initialised. Fine-tuning was then conducted by unfreezing the entire model and re-training it on the target dataset. This methodology is detailed further in the Fastai library documentation (Howard & Gugger, 2020).

## 3. RESULTS

All our tests were performed in Python using Jupyter Notebook on Google Collaboratory platform. It enabled us to use Linux machines to run the code, and high performance nVidia Tesla V100 GPU for computations. We used PyTorch version 1.3.1+cu100, and fast.ai version 1.0.59 libraries for neural network training. We achieved accuracy of 48% for the three-class system and 62% for the two-class classification system, as presented in Table 3.

**Table 3.** Summary of classification results for different CNN architectures on Datasets A1 & A2 (Fixations) and Dataset B (Scan paths)

| Architecture | Classes | Accuracy |
|---|---|---|
| ResNet-18 | (Dataset A1) C, D | 62.5% |
| ResNet-34 | (Dataset A1) C, D | 50% |
| ResNet-18 | (Dataset A1) A, C, D | 41.7% |
| ResNet-50 | (Dataset A1) A, C, D | 58.3% |
| ResNet-18 | (Dataset A2) C, D | 55% |
| ResNet-50 | (Dataset A2) C, D | 53.8% |
| ResNet-18 | (Dataset A2) A, C, D | 35% |
| ResNet-50 | (Dataset A2) A, C, D | 27.5% |
| ResNet-50 | (Dataset A2 448px) A, C, D | 48.3% |
| ResNet-18 | (Dataset B) C, D | 85.7% |
| ResNet-18 | (Dataset B) A, D | 50% |
| ResNet-18 | (Dataset B) A, C, D | 45% |
| ResNet-34 | (Dataset B) A, C, D | 54.5% |
| ResNet-50 | (Dataset B) A, C, D | 63.6% |
| ResNet-101 | (Dataset B) A, C, D | 45% |
| ResNet-151 | (Dataset B) A, C, D | 63.6% |

Note: Groups: C - Control, D - Depressive, A – Anxious

ResNet-18 network trained on Dataset B achieved the classification accuracy of 85.7%, but in the subsequent runs with a random train/test split that model achieved a much lower average accuracy of around 60%. The results indicate that our approach can differentiate between affective disorders and a control group. We also see that an increase in the number of the layers of ResNet does not automatically improve the results (e.g. simpler ResNet-18 and more complex ResNet-101 produce the same classification results).

Social anxiety and depression proved difficult to distinguish between each other. To investigate that further we repeated the ResNet-18 classification on randomised Dataset B five times (reduced to anxious and depressive participants) and received an average accuracy of 50%. The ResNet-50 systems have achieved consistently better classification results reaching 63.6%. They classified seven out of eleven people correctly, with only four people classified incorrectly (Figure 5).

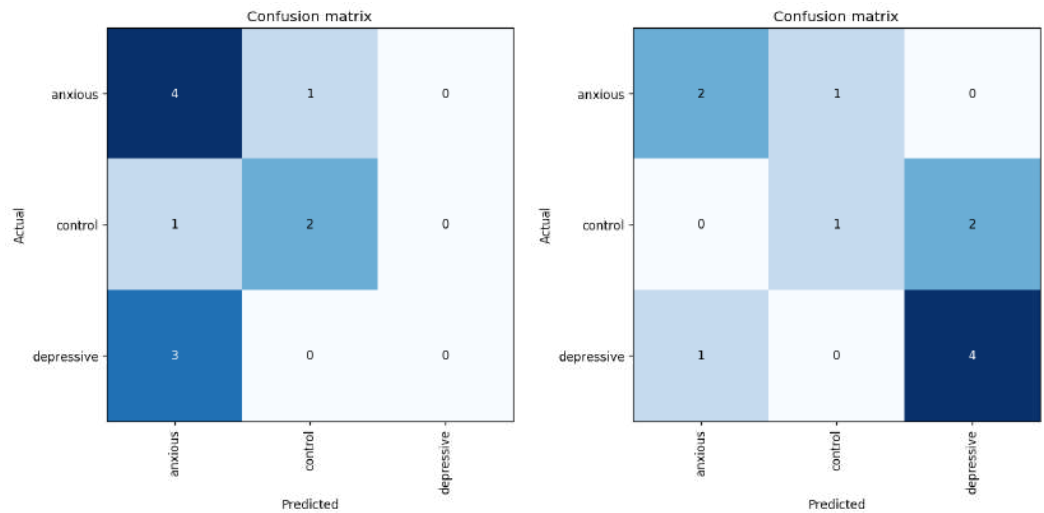

**Figure 5.** Confusion matrices presenting results for ResNet-34 system (Accuracy of 54.5%) and ResNet-50 (Accuracy of 63.6%) on Dataset B (Scan paths)

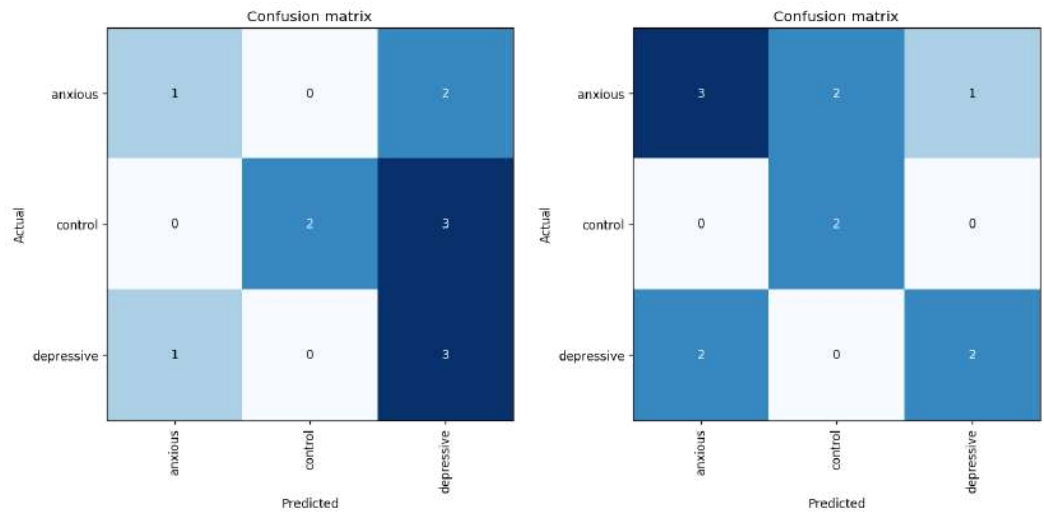

**Figure 6.** Confusion matrices presenting results for ResNet-34 system (Accuracy of 50%) and ResNet-50 system (Accuracy of 58.3%) on Dataset A1

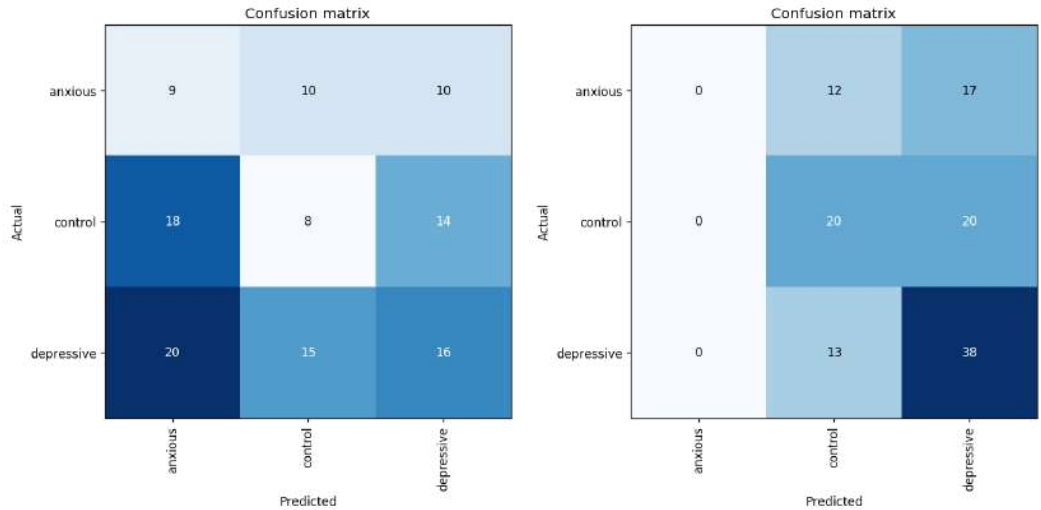

**Figure 7.** Confusion matrices presenting results for ResNet-50 system using image size of 224x224px (Accuracy of 27.5%) and 448x480px (Accuracy of 48.3%) on Dataset A2

The CNN systems using a much larger Dataset A2 (see Table 2) did not produce better results than those fed with images of clean, raw scan paths of Dataset B. Extracting gaze pattern proves to be the best way to reduce noise in deep learning approach and could be explored further. The selected confusion matrices for both configurations are presented in Figures 6 and 7.

## 4. DISCUSSION

The main contribution of the present paper is the first attempt to build a three-class computer-based system for screening of affective disorders, depression, and social anxiety, from gaze scan paths. This technique could complement traditional diagnostic methods, such as questionnaires and structured interviews. Unlike screenings for organic diseases, where false positives waste resources and cause anxiety, depression screening can identify subthreshold symptoms warranting follow-up (Zeng et al., 2024).

Recent studies have highlighted the potential of eye-tracking as a tool for identifying psychiatric disorders. For example, a recent eye-tracking study found that patients with depression exhibit distinct eye movement patterns compared to healthy controls, suggesting that these features could serve as biomarkers for depressive disorders. While Gao et al. (2023) propose eye movement recognition as a supplementary method for diagnosing depression, their study did not report the accuracy of classification models.

In another study, Kim et al. (2024) applied deep learning methods to classify psychosis and obsessive-compulsive disorder. Using a Long Short-Term Memory (LSTM) model implemented in Python with PyTorch, they achieved a binary classification accuracy of 80.7%. Their findings demonstrate that eye-tracking-based deep learning models can directly and rapidly identify impaired executive function during visuospatial memory encoding, underscoring the applicability of this approach across a range of psychiatric and neurological conditions.

By comparison to the LSTM model, our model achieved lower accuracies of 48% for a three-class system and 62% for a two-class system, primarily due to being trained on a smaller dataset. Nevertheless, these results support the feasibility of using eye-tracking data in AI-driven diagnostic tools. Future research should focus on expanding datasets and refining models to enhance classification accuracy and broaden the applicability of this technology to organisational and clinical settings.

The Choi et al. (2024) study focuses on utilizing digital phenotypes and feature representation learning to classify social anxiety disorder. Their study achieved impressive accuracy in predicting the severity of social anxiety symptoms, with an 87% accuracy for the primary symptoms, but uses completely different dataset based on multiple digital phenotypes like app usage, phone usage, call patterns, call logs, movement patterns, environmental patterns, and physiological patterns. Obviously, such a dataset gathered over a 7 to 13-week period will have more predictive power than our dataset, which is based solely on scan paths generated as a result of a free viewing task.

We tested five CNN architectures to choose the most suitable one. The results showed that ResNet-50 models achieved higher predication accuracy than ResNet-18 models for both depression and social anxiety classes. The method is fast, and computations related to training of our ResNet-150 model on small Dataset B using nVidia Tesla V100 GPU took 4 minutes only, whereas the average training time for ResNet-50 on a larger Dataset A2 was 15 minutes. Table 3 compares our method with other recent studies in the field.

AI-Based Screening for Depression and Social Anxiety Through Eye Tracking: An Exploratory Study.Table 4. Comparison of recent studies focusing on similar mental health screening systemsTable 4. Comparison of recent studies focusing on similar mental health screening systems

| Study | Year | Disorder | Methodology | Key Findings | Accuracy |
|---|---|---|---|---|---|
| Gao et al. | 2023 | Depression only | Statistical analysis of Eye-tracking data | Eye movement features differ in patients with depression, suggesting biomarkers | Accuracy not reported |
| Kim et al. | 2024 | Psychosis and OCD | LSTM model for binary classification | Eye-tracking deep learning models identify executive function impairment in psychosis and OCD | 80.7% (binary classification) |
| Abinaya & Vadivu | 2024 | Social Anxiety only | Machine learning, clustering, data exploration | Identified subgroups with social anxiety using SPIN questionnaire; suggests VR and AR for therapy | Not applicable (focus on clustering and machine learning) |
| Choi et al. | 2024 | Social Anxiety only | Digital phenotypes, feature representation learning | Achieved 87% accuracy for predicting social anxiety severity using mobile phones and seven digital phenotypes | 87% (binary classification) |
| Our Study | 2024 | Depression and Social Anxiety | CNN (ResNet) models for binary and three-class classification (Python, PyTorch) | Classifying depression and social anxiety using eye-tracking data (scan paths of visual attention) gathered during a 10-second free viewing task | 48% (three-class system), 62% (binary classification) |

Although the selected augmentation of the datasets did not visibly improve the efficiency of the models, the result suggests that if transfer learning is used, even a small sample of a few hundred images (in our case a 600 sample) can be sufficient to predict the distortion of typical attention biases from scan paths of visual attention.

## 5. CONCLUSION

The results of this study show that gaze-pattern classification is a promising method to quantify person's well-being through screening of affective disorders. Although the preliminary tests met our expectations, the approach can be improved further by generating better data using generative adversarial networks, and retraining the models on a larger, more balanced dataset. Adversarial networks, consisting of two neural networks—a generator and a discriminator—that are trained simultaneously in a competitive framework where the generator aims to create realistic data instances, and the discriminator attempts to distinguish between real and generated instances, are still state-of-the-art in image data generation (Goodfellow et al., 2020). Another option for future work is to re-evaluate the newer CNN architectures, that were proven useful in medical image classification (Iqbal, N. Qureshi, Li, & Mahmood, 2023).

To conclude, we demonstrated a proof-of-concept and its practical applicability for gaze-based classification of social anxiety and depression. We propose that this computer-based method, designed to predict attentional biases, could serve as a foundation for an innovative mental health screening system. In the future, such systems could be developed using affordable and accurate eye-trackers integrated into virtual reality headsets, enabling organisations to proactively monitor and support employee well-being while fostering healthier and more productive work environments.

## ACKNOWLEDGEMENTS

We thank Michael Connolly for proofreading.

**How to cite this article:**